\documentclass[journal]{IEEEtran}
\usepackage{amsmath,amsfonts}
\pdfoutput=1  
\usepackage{algorithmic}
\usepackage{array}
\usepackage{amssymb}
\usepackage{cite}
\usepackage{graphicx}  
\usepackage{hyperref}
\usepackage{url}
\usepackage{makecell}
\usepackage[caption=false,font=normalsize,labelfont=sf,textfont=sf]{subfig}  
\usepackage{textcomp}
\usepackage{stfloats}
\usepackage{verbatim}
\usepackage{multirow}
\usepackage{colortbl}  
\usepackage{balance}

\def\BibTeX{{\rm B\kern-.05em{\sc i\kern-.025em b}\kern-.08em
    T\kern-.1667em\lower.7ex\hbox{E}\kern-.125emX}}

\begin{document}
\title{Boundary and Position Information Mining for Aerial Small Object Detection}

\author{Rongxin Huang,Guangfeng Lin,Wenbo Zhou,Zhirong Li,Wenhuan Wu
\thanks{(Corresponding author: Guangfeng Lin.)
Rongxin Huang,Guangfeng Lin,Wenbo Zhou,Zhirong Li,
Xi'an University of Technology, Xi'an 710054,
China (e-mail: lgf78103@xaut.edu.cn); Wenhuan Wu, School of Electrical and Information Engineering, Hubei University of Automotive Technology, Shiyan 442002, China.
Copyright (c) 20xx IEEE. Personal use of this material is permitted.
However, permission to use this material for any other purposes must be
obtained from the IEEE by sending a request to pubs-permissions@ieee.org.
.}}
\markboth{Journal of \LaTeX\ Class Files,~Vol.~14, No.~8, August~2024}%
{Shell \MakeLowercase{\textit{et al.}}: Boundary and Position Information Mining for Aerial Small Object Detection}

\maketitle

\begin{abstract}
Unmanned Aerial Vehicle (UAV) applications have become increasingly prevalent in aerial photography and object recognition. However, there are major challenges to accurately capturing small targets in object detection due to the imbalanced scale and the blurred edges. To address these issues, boundary and position information mining (BPIM) framework is proposed for capturing object edge and location cues. The proposed BPIM includes position information guidance (PIG) module for obtaining location information, boundary information guidance (BIG) module for extracting object edge, cross scale fusion (CSF) module for gradually assembling the shallow layer image feature, three feature fusion (TFF) module for progressively combining position and boundary information, and adaptive weight fusion (AWF) module for flexibly merging the deep layer semantic feature. Therefore, BPIM can integrate boundary, position, and scale information in image for small object detection using attention mechanisms and cross-scale feature fusion strategies. Furthermore, BPIM not only improves the discrimination of the contextual feature by adaptive weight fusion with boundary, but also enhances small object perceptions by cross-scale position fusion. On the VisDrone2021, DOTA1.0, and WiderPerson datasets, experimental results show the better performances of BPIM compared to the baseline Yolov5-P2, and obtains the promising performance in the state-of-the-art methods with comparable computation load.
\end{abstract}

\begin{IEEEkeywords}
UAV, small object detection, YOLOv5, adaptive weight fusion, boundary information, cross-scale feature fusion.
\end{IEEEkeywords}

\section{Introduction}
\IEEEPARstart{U}{nmanned} Aerial Vehicle (UAV) technology has been widely utilized across various fields, which include intelligent transportation, disaster relief, agriculture, and human flow monitoring. One of the key research areas within these applications is object detection in aerial photography by UAVs. UAV aerial images usually have complex backgrounds, small target sizes, and motion-induced blur, which exacerbate the difficulty for effectively capturing small target features. The object detection faces significant challenges for small target perception in the specific situations, which include extreme scale imbalance among detected targets, a wide range of target size distributions, and a high proportion of small-sized targets in aerial photography. Extreme scale imbalance means that the small targets occupy very few pixels compared to large targets, resulting in texture and shape losses due to different attention of target scale in the complex background. The wide range of target size distribution stands for that targets have diverse sizes with varying distances in aerial photography. The high proportion of small-sized targets needs the high density perception for greater accuracy of object detection.

\begin{figure}[t]
\centering
\includegraphics[width=1\linewidth]{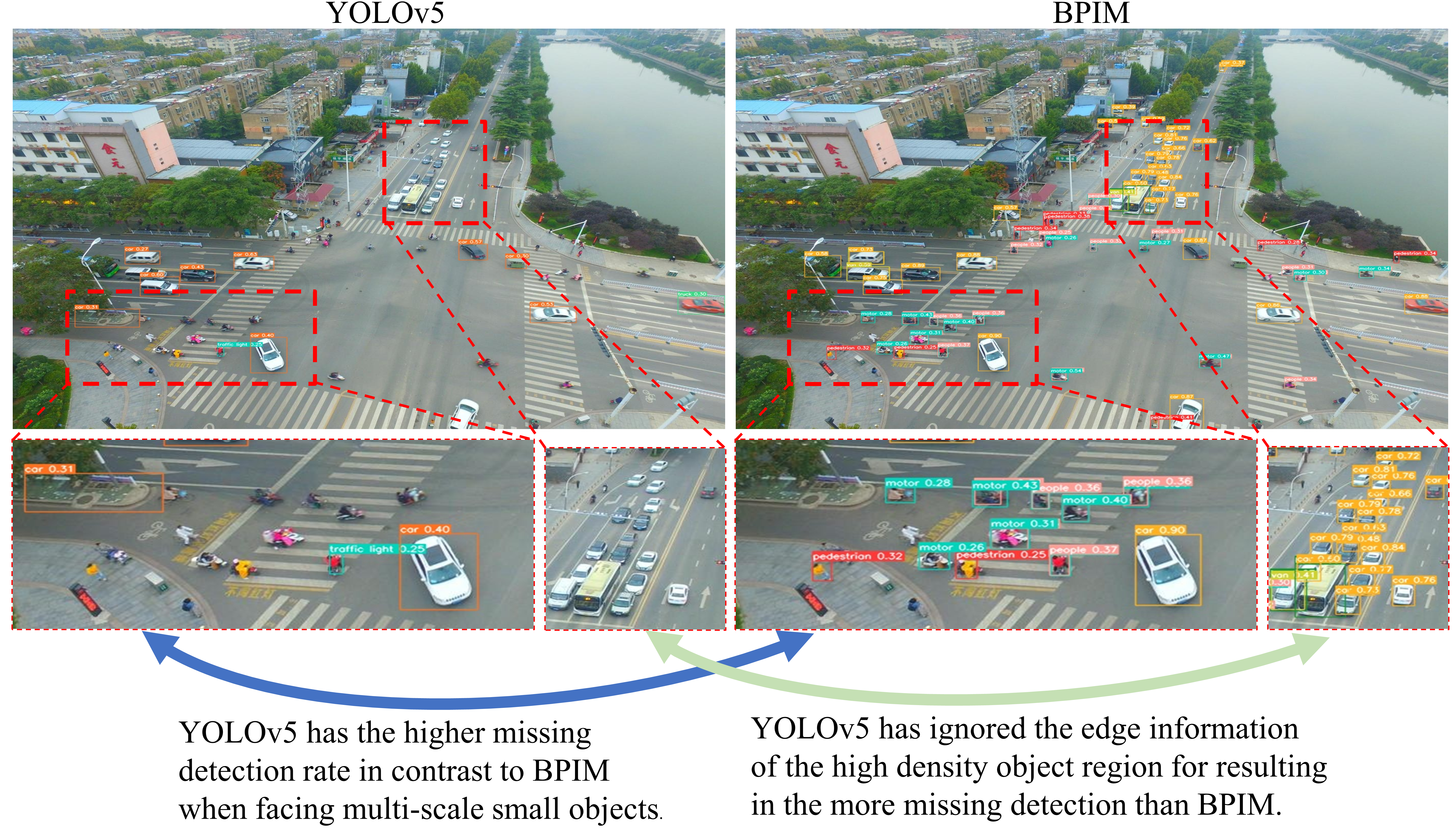}
\caption{\raggedright The limitations of existing models in comparison with BPIM.}
\label{fig:1}
\end{figure}

Existing object detection methods primarily deal with the medium or large size targets and often struggle in numerous small targets. In addition, these detection methods tend to lose small target features as network depth increases, resulting in missed detections and false alarms. To address the inadequate attention on small targets in existing methods, various methods have used the attention mechanisms \cite{ref1, ref2, ref3, ref4, ref5, ref6} and the multi-scale fusion strategies \cite{ref7, ref8, ref9, ref10, ref11} for enhancing small target detection through limited information mining in images. However, existing attention mechanisms often fail to focus on the adaptive integration of the boundary features and location information of small targets, limiting detection capabilities. Moreover, traditional multi-scale feature fusion strategies often overlook the inconsistency of information across different scales, reducing detection accuracy. The limitations of existing models in comparison with BPIM are shown in Figure~\ref{fig:1}.

The proposed BPIM is a small target detection framework for aerial photography based on YOLOv5 \cite{ref12} to address the above issues. BPIM aims to improve detection accuracy and robustness for small targets by cross-scale feature fusion and a new attention mechanism. Furthermore, BPIM dynamically adjusts the weights to address the information inconsistency of contextual features by adaptive weight fusion with boundary and position information. The proposed BPIM includes a position information guidance (PIG) module for obtaining  location information, a boundary information guidance (BIG) module for extracting object edges, a cross scale fusion (CSF) module for gradually assembling the shallow layer image features, a three feature fusion (TFF) module for progressively combining position and boundary information, and an adaptive weight fusion (AWF) module for flexibly merging the deep layer semantic features. The integration architecture of these modules  is referred to as BPIM in Figure~\ref{fig:2}. The main contributions are as follows.
\begin{enumerate}
    \item The adaptive weight fusion with boundary can capture small target boundaries and position features by the different scale edge information extraction and the feature mapping weight adaptively computed. Moreover, this strategy can obtain low-level appearance features through the shallow multi-scale edge fusion structure, and also obtain high-level semantic features by the deep contextual information fusion structure in the different depth layers.
    \item The cross-scale feature interaction can excavate whole -image features by the spatial self-attention computation and 3D convolution transform. This way complements the boundary and position features with the whole image feature for enhancing the small object location perception through the link between shallow and deep layers.
    \item The promising results of small object detection can demonstrate the efficiency of the proposed BPIM across different datasets. Furthermore, the proposed BPIM offers better mean Average Precision (mAP) values for accurate object perception and a smaller increase in parameter numbers for computational efficiency compared to existing state-of-the-art methods.
\end{enumerate}

\section{Related Work}

The definition of small objects can be divided into two ways based on object size. One is the size based on relative scale, which calculates the ratio of a target's bounding box area to the total image area. Relative sizes between 0.08\% and 0.58\% are considered small. The other is based on absolute scale, which is described by image resolution.For instance, the MS-COCO dataset \cite{ref29} defines small targets as having a resolution of less than 32 × 32 pixels. Similarly, there are different definitions of small targets across various aerial imagery datasets. For example, the DOTA dataset \cite{ref27} clarifies small targets as those with pixel values between 10 and 50, while the TinyPerson dataset \cite{ref28} defines them as small targets with pixel values between 20 and 32. Existing small object detections mainly focus on feature fusion, attention mechanisms, and YOLO series methods to mine the limitation information of small objects. Therefore, the related works review three points as follows.

\subsection{Feature Fusion for Object Detection}

One of the main challenges in small target detection is the gradual weakening of feature information as the depth of the neural network increases. Multi-scale feature fusion is an effective technique to enhance small target detection to mitigate this issue. Multi-scale feature fusion in existing methods image pyramids and feature pyramids.

Image pyramids can obtain features from different-scale images for capturing object information by the high-resolution image pyramid network\cite{ref9}, and different parameter sizes processing different resolution levels of the image pyramid\cite{ref37}. Because image pyramids only focus on the low-level appearance information of pixel values, while neglecting high-level semantic information mining. Feature pyramids can further find the information dependence among various scale features for enhancing the object perception by the top-down feature pyramid \cite{ref7}, vector learning for deep feature clarity\cite{ref8}, multi-scale feature sets alignment and fusion\cite{ref9}, the multi-component fusion network\cite{ref10}, the recursive pyramid network\cite{ref11}, the dynamic feature pyramid networks\cite{ref38}, the augmented weighted bidirectional feature pyramid network\cite{ref39}, and multi-level feature pyramid transformer\cite{ref40}. However, existing multi-scale fusion methods often overlook the information inconsistency when fusing features across different scales, leading to higher rates of missed detections.


\subsection{Attention Mechanism for Object Detection}

The other challenges in small object detection is insufficient attention on small targets. Therefore, the various attention mechanisms enhance the small target characterization capabilities of neural networks to better perceive the small-sized instances in the scene. Existing attention mechanisms include two types of methods, which are interactions in different dimensions and attention maps based on the spatial information.

The interactions in different dimensions can focus more on small targets through channel and spatial dimensions\cite{ref3}, interplays based on global features while reducing redundancy\cite{ref5}, a triple-branching structure to capture cross-dimensional interactions and compute attention weights\cite{ref6}, a Transformer architecture integrated into the detector's prediction head and incorporating CBAM into the network's neck portion\cite{ref30}, and soft-pooling and a subspace attention module to address the loss of critical edge information in small targets\cite{ref1}. The interactions of different dimensions primarily aim to fully obtain the dimensional information of the image, while attention maps based on the spatial information focus on the important parts of the image. These attention map methods improve object perception by the coordinate attention mechanism to create a directional location-aware attention map\cite{ref4}, separate network branches that stack ViT blocks to effectively harness the spatial transformation potential of ViTs in a divide-and-conquer fashion\cite{ref41}, and an attention-based encoder on ViT for salient object detection to ensure the globalization of representations from shallow to deep layers\cite{ref42}. However, these methods often fail to simultaneously consider the edge and position information of small targets, and do not fully exploit the limited information available for small targets.

\subsection{YOLO Series method for Object Detection}

Single-stage detection algorithms like the YOLO \cite{ref16} and SSD \cite{ref17} series can directly classify targets across all spatial regions of an image. These algorithms detect targets with a straightforward structure, offering faster speeds and lower computational demands, which makes them advantageous for real-time object detection in UAV applications.

YOLOv5 \cite{ref12}, the first PyTorch-based version of YOLO, has a smaller model size, faster speeds, and lower memory usage compared to YOLOv4 \cite{ref18}. YOLOv7 \cite{ref20} introduces the extended efficient layer aggregation network (E-ELAN) and an auxiliary head to optimize model architecture and training, making it compatible with the embedding devices. However, its complex architecture demands the significant computational resources and performs sub-optimally in detecting small targets and dense scenes. YOLOv8 \cite{ref21} supports various visual tasks that are both scalable and suitable for engineering applications. It boasts high flexibility and detection efficiency, with good generalization ability on large datasets. However, its complexity also necessitates considerable computational resources and detection time. YOLOv10 \cite{ref22} adopts an efficiency-accuracy-driven model design strategy, optimizing YOLO components from both efficiency and accuracy perspectives, significantly reducing computational overhead while improving performance.

While the YOLO family is known for fast detection speed, high accuracy, strong generalization capability, and ease of deployment, the continually improving YOLO algorithm can better adapt to small target detection scenarios. Drone-YOLO \cite{ref23} can enhance small target detection by using a three-layer PAFPN structure and incorporating a small target detection head with a large-scale feature map. AS-YOLOv5 \cite{ref1} can address the loss of critical edge information for small targets by enhancing the feature extraction network with soft-pooling. It employs a feature fusion method with learnable parameters to effectively rebalance feature layers, ensuring that small target information remains prominent and is not overshadowed by larger target features.

Considering both accuracy and speed for small object detection, we choose the user-friendly YOLOv5 based on the PyTorch framework. YOLOv5 is easier to train, deploy, and further optimize, making it an ideal choice for our purposes.

\begin{figure*}[ht]
\centering
\includegraphics[width=1\linewidth]{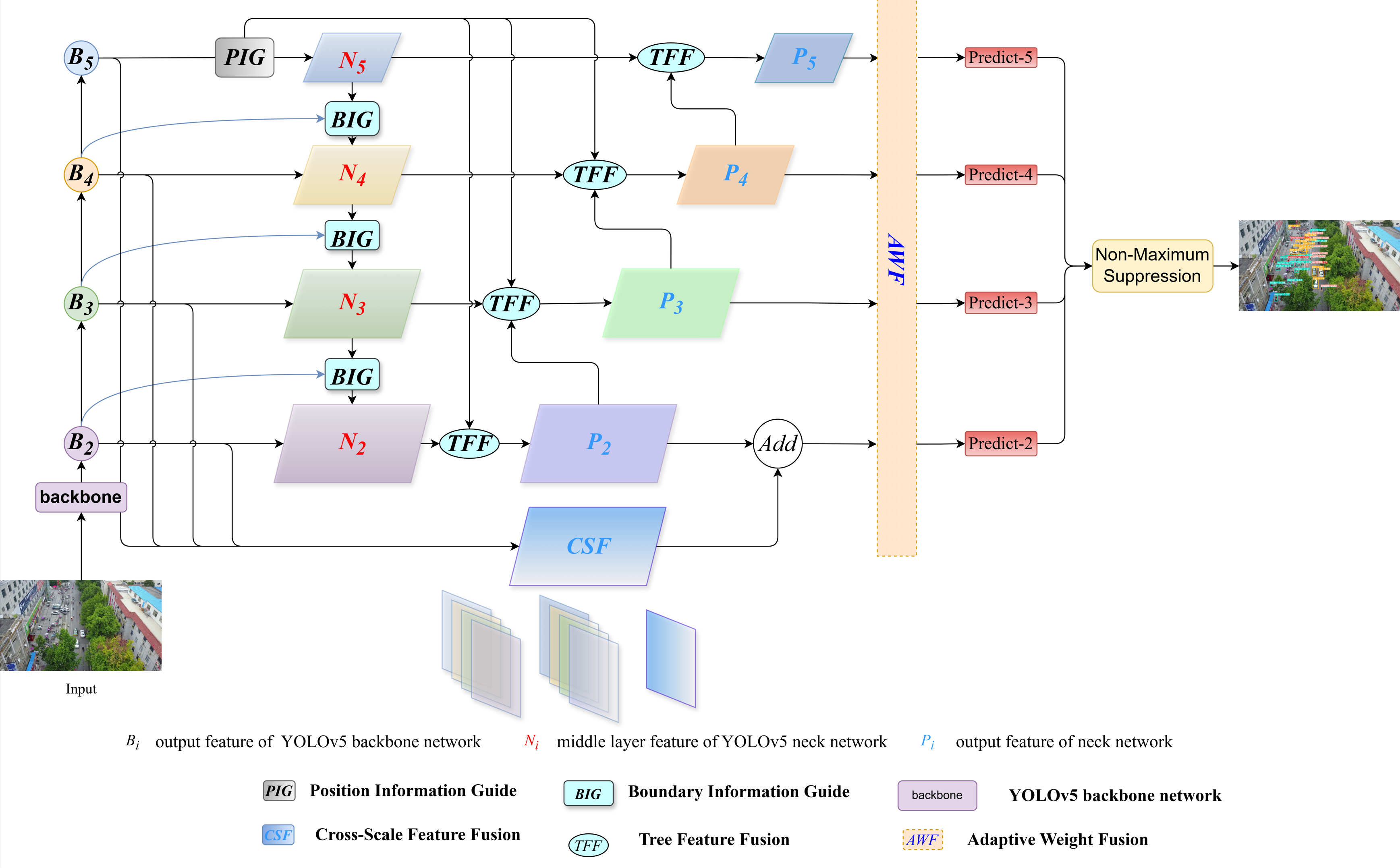}
\caption{\raggedright Overall structure of our framework BPIM}
\label{fig:2}
\end{figure*}

\section{Method}
\subsection{Model Architecture}
To accommodate different scenarios, YOLOv5 offers five models with varying depths and widths: YOLOv5s, YOLOv5m, YOLOv5l, YOLOv5x, and YOLOv5n\cite{ref12}. The primary differences among these models lie in their size and computational complexity. YOLOv5s is the smallest version, featuring the fewest layers and the least computational complexity, resulting in the lowest detection performance. In contrast, YOLOv5x is the largest version, with the most layers and highest computational complexity, and thus has the best detection performance. YOLOv5n and YOLOv5l strike a better balance between detection speed and accuracy, making them suitable for aerial image object detection. Therefore, we select YOLOv5n and YOLOv5l as baseline models and add a small object detection layer to enhance small target detection capabilities. Our goal is twofold: one improves the algorithm's efficiency in utilizing multi-scale feature information and mining contextual information, the other is to enhance the model's ability to extract and perceive edge and position information from small targets, thereby boosting the network's small target detection performance. The specific model architecture, illustrated in Figure~\ref{fig:2}, includes position information guidance (PIG) module for obtaining location information, boundary information guidance (BIG) module for extracting object edge, cross scale fusion (CSF) module for gradually assembling the shallow layer image feature, three feature fusion (TFF) module for progressively combining position and boundary information, and adaptive weight fusion (AWF) module for flexibly merging the deep layer semantic feature.

\subsection{Adaptive weight fusion with boundary}
To address the limitations that existing algorithms have in exploiting contextual boundary and multi-scale feature information for small targets, we designed BIG and AWF modules. The AWF modules for the adaptive weight fusion of the different scale information with the BIG module for edge information mining can enhance the efficiency of small object detection in aerial images.
\begin{figure}
\centering
\includegraphics[width=1\linewidth]{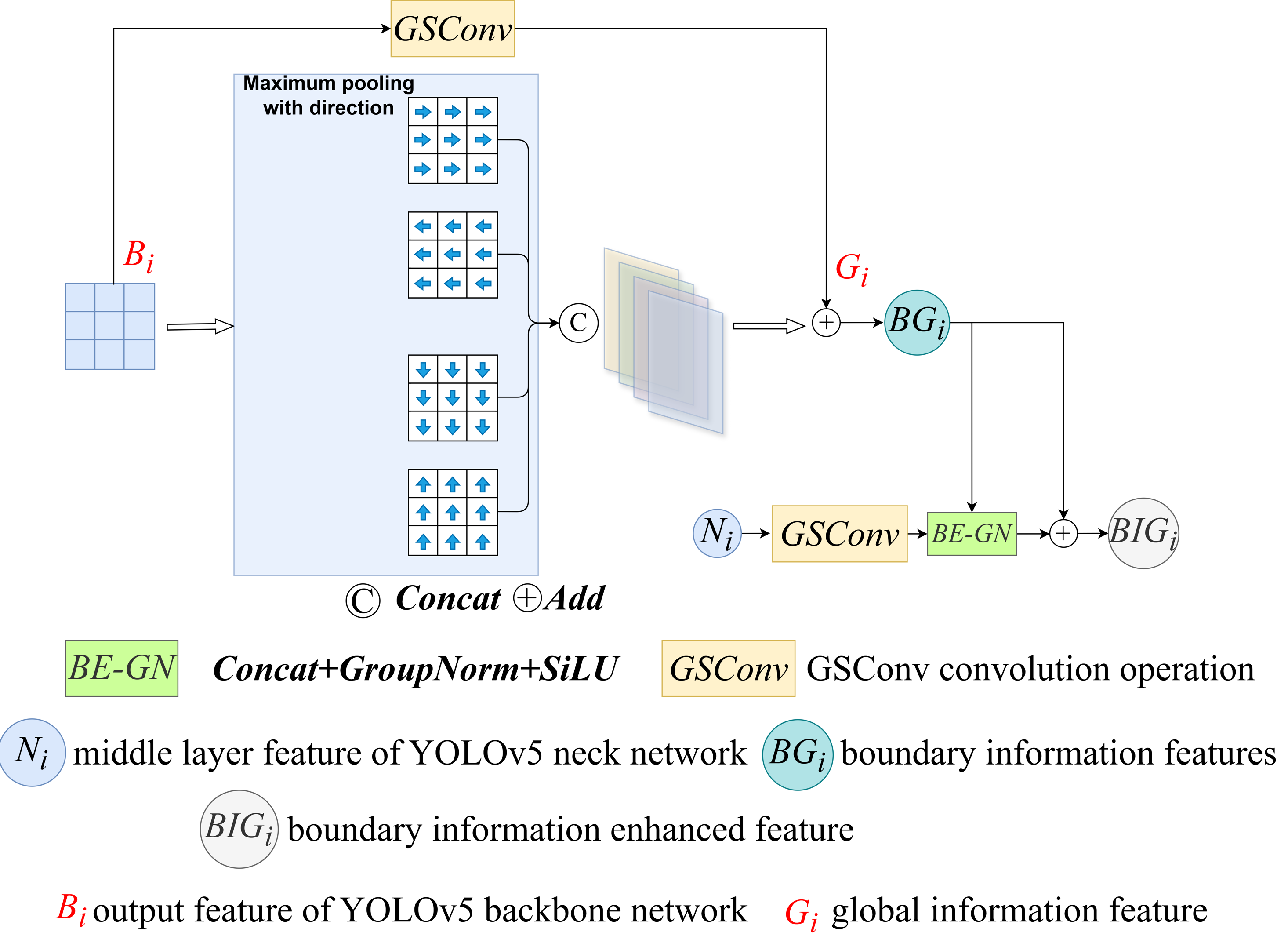}
\caption{\raggedright Boundary Information Guidance(BIG) Module}
\label{fig:5}
\end{figure}
the BIG module can boost the detection of dense small targets by leveraging both contextual boundary feature information. The BIG module uses boundary-enhanced features, which emphasize the edges of small targets, to effectively mine the contextual information of edge in different scales. The BIG module is shown in Figure~\ref{fig:5}. To ensure effective utilization of shallow features in the rich details and boundary information,the input features of the BIG module include the neck network feature $N_i$ from the middle part of YOLOv5 and the features $B_i$ from the backbone network of YOLOv5. The output of the BIG module is denoted as $BIG_i$, $i=2,3,4,5$ is the number of the different layers. The contextual information of boundary can be obtained by the cross-cascading of the BIG module and the neck network at different scales in Figure~\ref{fig:2}. The BIG module shows in Eq. ~\ref{eq:3} for getting the interaction information between the middle feature of YOLOv5 neck network and the boundary information. The output of the BIG module is $BIG_i$ as follows.

\begin{equation}
\begin{aligned}
&BIG_i=\\
& Add(BG_i, SiLU(GroupNorm(Concat(BG_i, GSConv(N_i)))))
\label{eq:3}
\end{aligned}
\end{equation}

$Add(\cdot)$ denotes the additive feature splicing operation. $SiLU(\cdot)$ denotes the activation function SiLU\cite{ref44}. $Concat(\cdot)$ is the concatenation operation. $GroupNorm$ stands for the group normalization\cite{ref43}. $GSConv(\cdot)$ is the GSConv of $1\times 1$ kernels to adjust dimensions. $BG_i$ can enhance the boundary information as follows.

\begin{equation}
BG_i = \text{Add}(G_i,\text{Boundary}(x))
\label{eq:4}
\end{equation}

$G_i$ is the global features by GSConv\cite{ref24} for meeting real-time requirements and reduces the number of model parameters in aerial small object detection. The Boundary($\cdot$) indicates the combination of the maximum pooling to capture boundary information from the different directions $l=\{left,right,top,bottom\}$ of the feature map in Eq.~\ref{eq:5} and Eq.~\ref{eq:51}.

\begin{equation}
\text{Boundary}(x)^l =
\begin{cases}
X_{i,j}^l, & j=W \\
\max \{X_{i,j}^l, X_{i,(j+1)}^l, \ldots, X_{i,W}^l\}, & j \neq W
\end{cases}
\label{eq:5}
\end{equation}

$X_{i,j}^l$ represents the feature map of the different direct $l$ at position ($i,j$), $W$ is the width value of the feature mapping. The boundary from the different directions is captured by maximum pooling based on the feature map traversing from the position ($i,j$) according to $l$ to detect abrupt changes in pixel values. Boundary$(x)$ is the integration of the different direction boundaries Boundary$(x)^l$ as follows.

\begin{equation}
\text{Boundary}(x) =Concat(\text{Boundary}(x)^l)
\label{eq:51}
\end{equation}

The above configuration ensures that the network effectively leverages the detailed boundary information from the shallow layers to enhance small target detection, especially in dense scenes. By integrating these components, the network achieves more robust and accurate detection performance.

\begin{figure}[ht]
\centering
\includegraphics[width=1\linewidth]{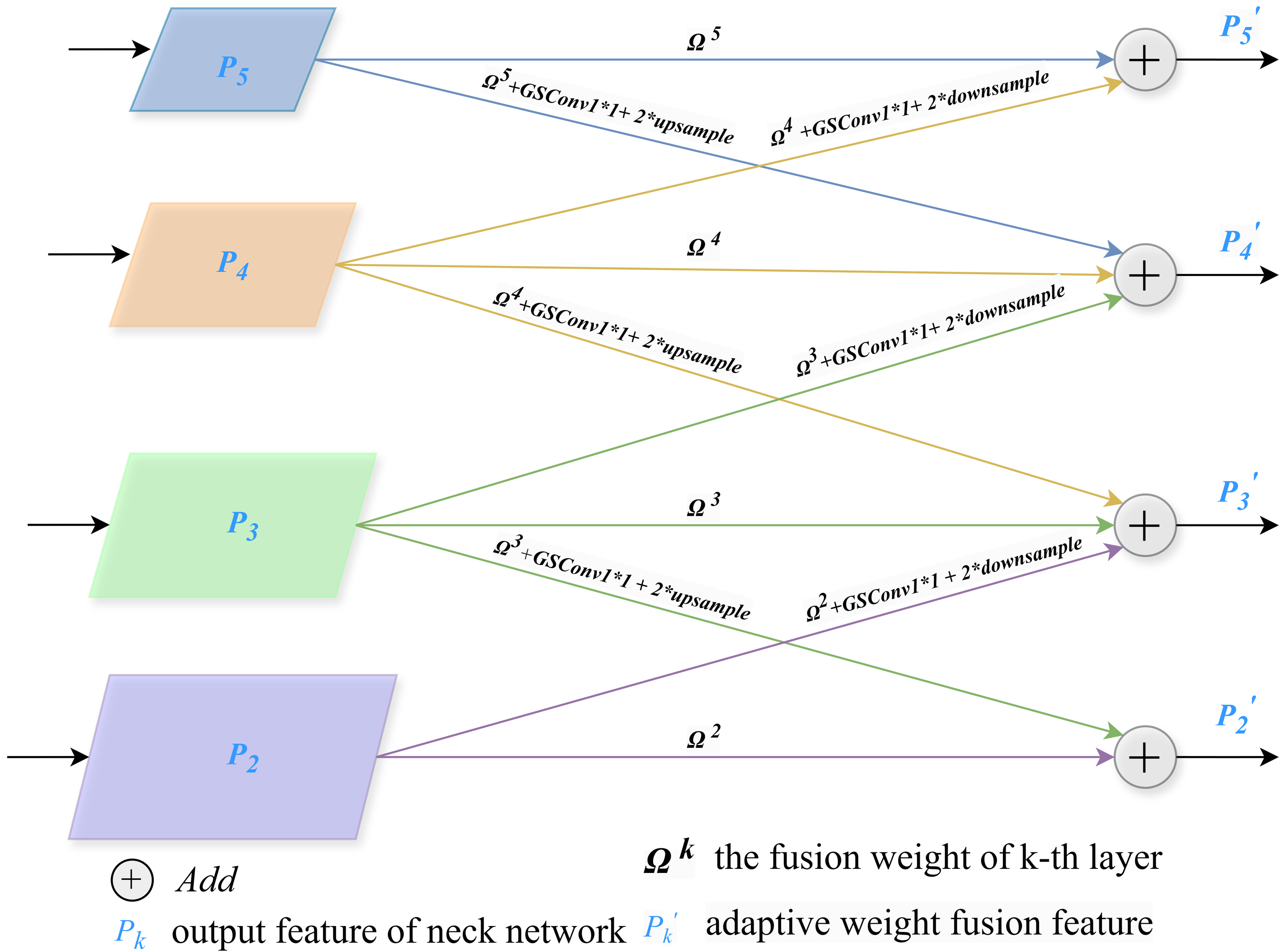}
\caption{\raggedright Adaptive Weight Fusion(AWF) module}
\label{fig:4}
\end{figure}

Furthermore, we propose the AWF module to improve multi-scale feature fusion efficiency while avoiding information redundancy based on the edge mining of the BIG module. The AWF module can effectively utilize feature information from different scales to enhance the detection head's fusion efficiency, as shown in Figure~\ref{fig:4}. Our method focuses on feature information interaction between adjacent layers in Eq.~\ref{eq:1}.

\begin{equation}
P_k' =
\begin{cases}
\Omega^k \bigodot P_k + \Omega^{k+1} \bigodot P_{k+1}, &k=2 \\
\Omega^{k-1} \bigodot P_{k-1} + \Omega^k \bigodot P_k + \Omega^{k+1} \bigodot P_{k+1}, &3 \leq k \leq 4 \\
\Omega^{k-1} \bigodot P_{k-1} + \Omega^k \bigodot P_k, &k=5
\end{cases}
\label{eq:1}
\end{equation}

\begin{equation}
\Omega^k =[w_{ij}^k]
\label{eq:1_1}
\end{equation}

$P_k$ represents the input features of the AWF moudle. $\bigodot$ is Hadamard product. $\Omega^k$ denotes the fusion weight contributed by the pixels on the feature maps of neck network $k$th layer. $\Omega^k$ is a tensor composed of the weights $w_{ij}^k$ of each pixel position. Eq. ~\ref{eq:2} shows $w_{ij}^k$ as follows.

\begin{equation}
w_{ij}^k = \frac{Conv_{1\times1}(e^{\lambda_{ij}^k})}{\sum Conv_{1\times1}(e^{\lambda_{ij}^k})}, \quad 0 \leq w_{ij}^k \leq 1
\label{eq:2}
\end{equation}

$\lambda_{ij}^k$ is the pixel of position $(i,j)$. $Conv_{1\times1}$ is 1×1 convolution. The adaptive weight fusion feature $P_k'$ can flexibly merge the deep layer semantic feature. This approach allows the network to learn the contribution of each pixel on the fused feature maps across different layers, rather than directly selecting a specific feature layer and discarding contributions from other layers as done in PANet. This allows BPIM to achieve better performance when facing small object detection.

\subsection{Cross-scale position fusion}
Attention mechanisms have been widely applied to enhance feature extraction in small object detection. However, existing methods primarily focus on single-scale feature extraction, ignoring cross-scale position information and its interrelationship. Therefore, we utilize PIG module and CSF module to enhance position information extraction ability of our framework. The shallow layers of a model network contain detailed information.Thus, we aim to comprehensively extract this hidden information from the shallow backbone network. To achieve this, we propose the CSF module, which uses 3D convolution to extract interaction information across the scale sequence direction for each scale.Furthermore, PIG module is designed to enhance the model's position awareness of target instances. Additionally, the Transformer coding layer is added to the final layer of the backbone network to integrate positional and intra-scale information. The output features from both the CSF and PIG modules are combined using the TFF module. This approach ensures efficient and rational utilization of features, sufficiently integrating cross-scale interaction and location-aware features.

\begin{figure}[hb]
\centering
\includegraphics[width=1\linewidth]{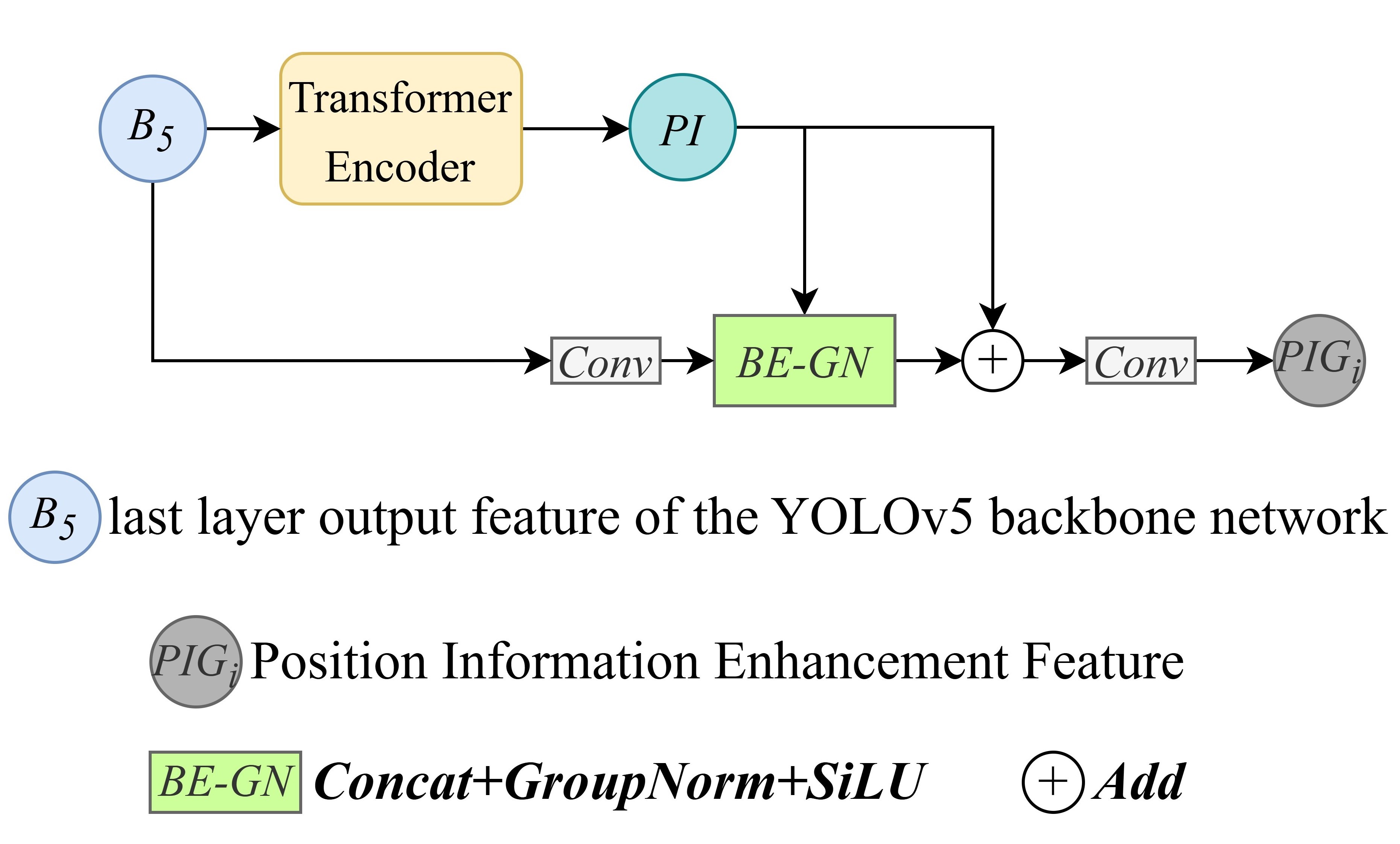}
\caption{\raggedright Position Information Guidance(PIG) Module}
\label{fig:7}
\end{figure}
The PIG module is shown in Figure~\ref{fig:7}, and introduces the multi-head self-attention mechanism using a single-layer Transformer encoder network. This mechanism extracts in-scale interaction and location information from the output $B_5$ of the last layer of the backbone network. PIG can learn the spatial relationships between pixels in the image and the long-range dependencies within the same scale, thereby enhancing the model's ability to perceive location information. The multi-head self-attention operates with three inputs: Query (Q), Key (K), and Value (V). The input data is divided into multiple heads, with each head processing a subset of the data. K is weighted and summed based on its similarity to Q. These weights can scale V. This process allows the model to discern differences between feature information at various locations and achieve a more robust feature representation. This mechanism is shown in Eq.~\ref{eq:6} as follows.
\begin{equation}
\text{MultiHead}(Q,K,V)=Concat(H_1,...,H_h)W^O
\label{eq:6}
\end{equation}

$h$ is the number of heads, $H_i$ is the output of the $i$-th head, and $W^O$ is the transformation matrix. The output of each head can be represented by Eq.~\ref{eq:7}.
\begin{equation}
H_i=\text{Attention}(QW_i^Q,KW_i^K,VW_i^V)
\label{eq:7}
\end{equation}
$W_i^Q$, $W_i^K$, and $W_i^V$ represent the transformation matrices for Query (Q), Key (K), and Value (V) of the $i$-th head. $\sqrt{d_k}$ is the scale factor. The attention calculation function, denoted as Attention($\cdot$), is typically implemented using the self-attention mechanism in Eq.~\ref{eq:8}.

\begin{equation}
\text{Attention}(Q,K,V)=\text{Softmax}\left(\frac{QK^T}{\sqrt{d_k}}\right)V
\label{eq:8}
\end{equation}

We adopt an 8-head self-attention model and set the dimension of the middle layer of the feed-forward network to 1024, using GELU\cite{gelu} as the activation function. The output $B_5$ from the last layer of the backbone network is converted from a tensor of [C,H,W] to [H×W,C] using the Flatten function, as shown in Eq.~\ref{eq:9}. After applying multi-head self-attention, we perform the inverse Flatten operation to revert the output features back to the same dimensions as $B_5$. This process results in the feature $PI$, which contains inter-scale interaction and positional relationship information. These features are further refined and enhanced using the boundary information enhancement method similar to BIG, so that the feature information with inter-scale location dependence is obtained. Finally, the features are scaled using different convolution steps to obtain scaled features $PIG_i$ ($i=2,3,4,5$) that match the size of the output features $N_i$ from the middle layers of the neck network, as shown in Eqs.~\ref{eq:10} and \ref{eq:11}.

\begin{equation}
Q=K=V=\text{Flatten}(B_5)
\label{eq:9}
\end{equation}

\begin{equation}
PI=\text{Reshape}(\text{MultiHead}(Q,K,V))
\label{eq:10}
\end{equation}

\begin{equation}
PIG_i=Conv(Add(PI,SiLU(GN(Concat(PI,Conv(B_5))))))
\label{eq:11}
\end{equation}

\begin{figure}[ht]
\centering
\includegraphics[width=1\linewidth]{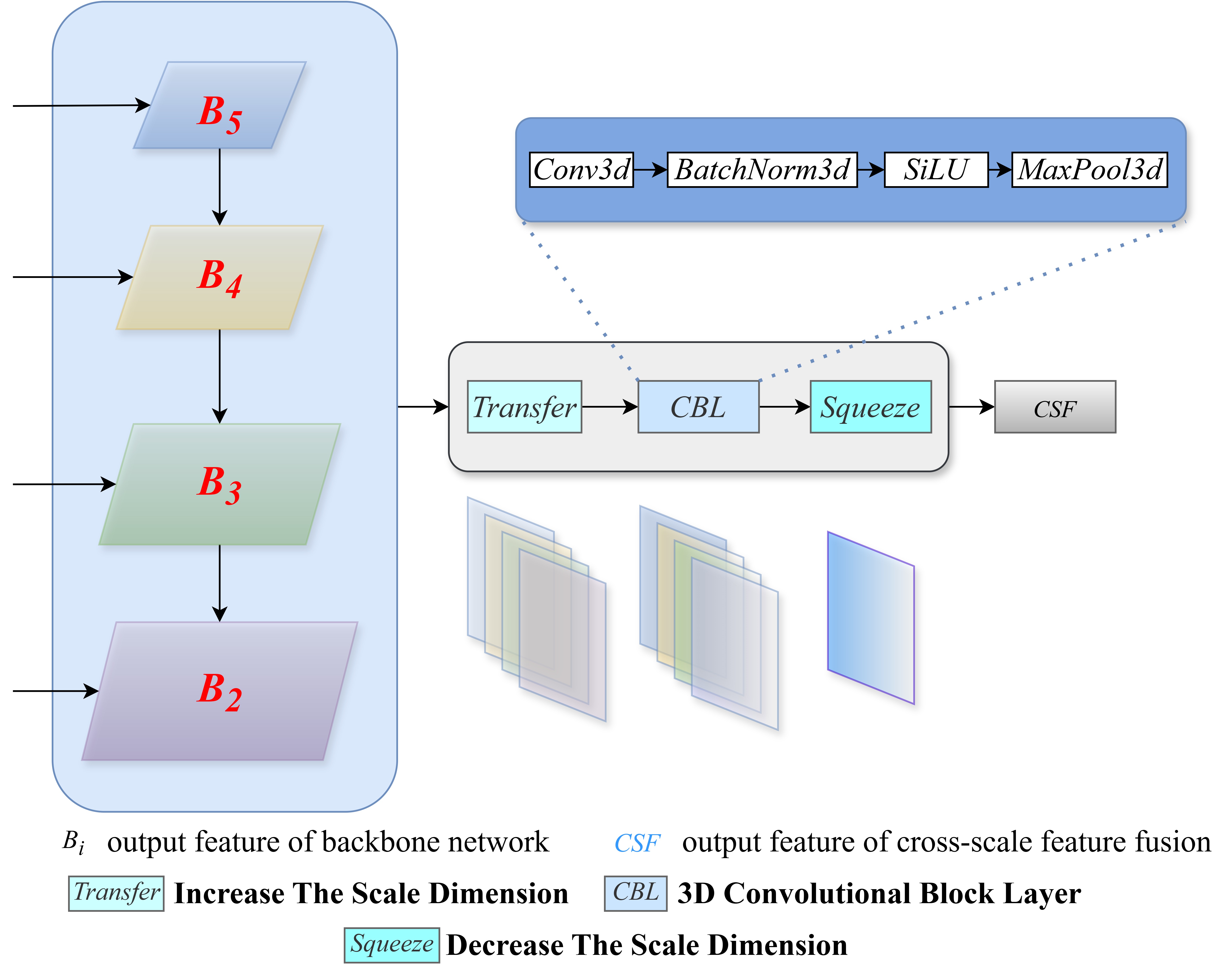}
\caption{\raggedright Cross-Scale Fusion(CSF) module}
\label{fig:8}
\end{figure}

3D convolution is effective for extracting temporal and spatial correlation information from videos along the time axis. In our approach, we use 3D convolution for feature extraction along the scale change direction. This enables us to capture cross-scale correlation information between feature maps at various scales, thereby enhancing the model's ability to represent target instances. To achieve this, we propose CSF module to extract cross-scale interaction information from each feature layer. The detailed structure of this module is depicted in Figure~\ref{fig:8}.

The output layer features of the backbone network contain richer detail information compared to the neck network, which is crucial for representing small target instances. Therefore, we use the output features $B_i$ ($i=2,3,4,5$) from the backbone network as input for the CSF module. As illustrated in Figure~\ref{fig:8}, the input feature $B_i$ ($i=2,3,4,5$) undergoes three operations: $Transfer$, $CBL$ and $Squeeze$ to obtain an output feature $CSF$ with cross-scale interaction information. The detailed process is as follows. Firstly, we use the $Transfer$ operation with unsqueeze function to adjust the input features $B_i$ of each scale to align the dimensions of all scale features. Secondly, it adds a scale direction dimension to each layer feature and fuses the features of each layer along this dimension. This process results in a cross-scale feature sequence $CS$, resembling a time series. Thirdly, the 3D Convolutional Block Layer ($CBL$) is used for cross-scale feature extraction, and the Squeeze function performs the inverse operation of $Transfer$, removing the scale direction dimension. In the end, we get output $CSF$ that incorporates cross-scale information in Eq.~\ref{eq:12} as follows.

\begin{equation}
CSF=Squeeze(CBL(CS))
\label{eq:12}
\end{equation}

$CBL(\cdot)$ performs several operations in sequence: 3D convolution, 3D batch normalization, SiLU activation and 3D maxpooling. As shown in Eq.~\ref{eq:13}, we use $Transfer(\cdot)$ to adjust the size of the dimensionally expanded input feature $B_i$ ($i=2,3,4,5$) to match the size of the output feature in the neck network. 
\begin{equation}
CS=Transfer(unsqueeze(B_2,B_3,B_4,B_5))
\label{eq:13}
\end{equation}

\begin{equation}
CSF=NearestNeighbor(CS)
\label{eq:14}
\end{equation}
$NearestNeighbor$ is nearest-neighbor interpolation for upsampling.

\begin{figure}[h]
\centering
\includegraphics[width=1\linewidth]{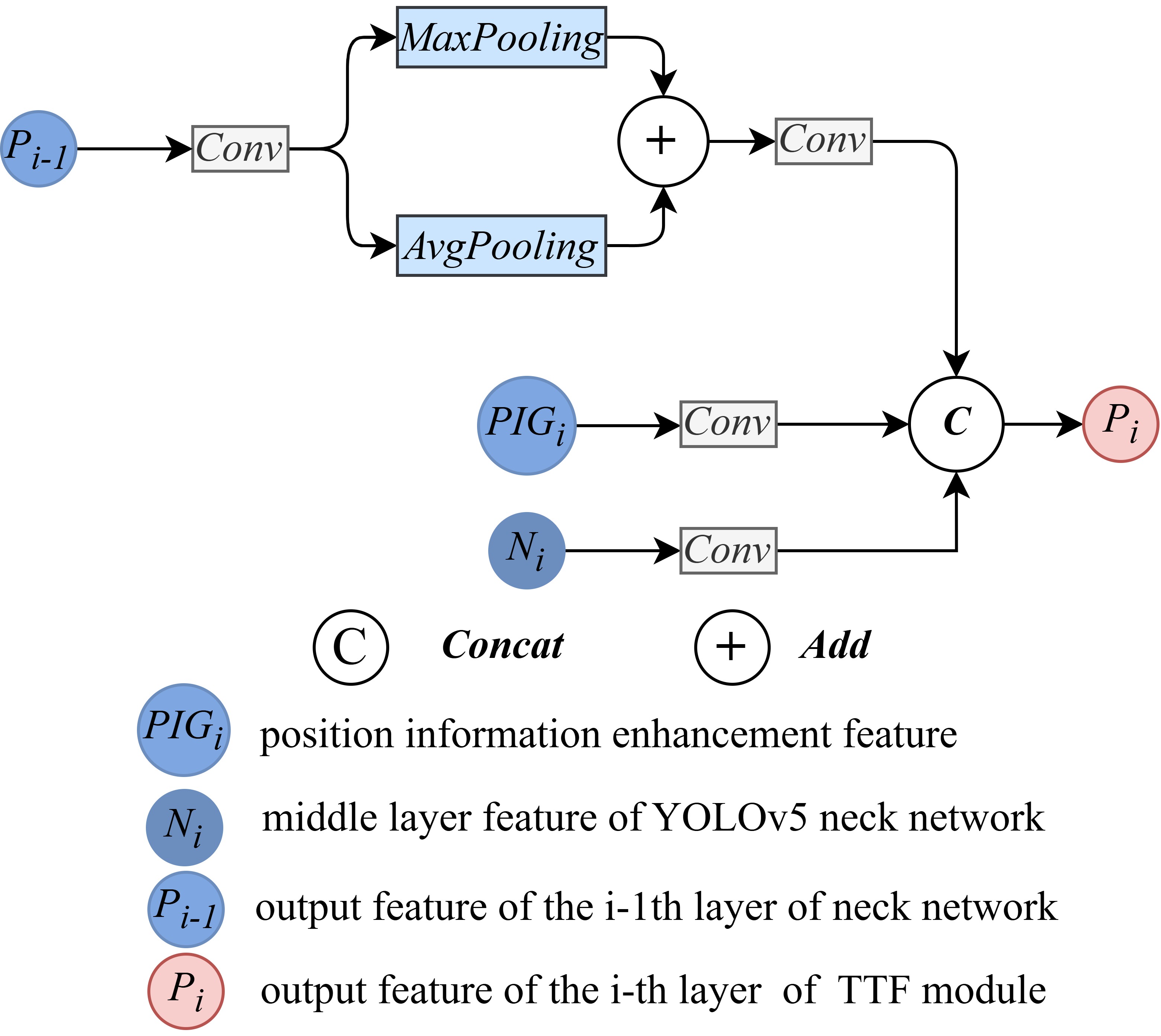}
\caption{\raggedright Three Feature Fusion(TFF) strategy}
\label{fig:9}
\end{figure}

In addition to rationally utilizing cross-scale information interaction features and location-aware features, we propose the TFF module for the output features of the CSF and PIG modules. In traditional PANet networks, feature maps are usually resized by upsampling small-scale features or downsampling large-scale features. However, downsampling large-scale features with a simple convolution operation often results in the loss of rich detail information in these features. To address this issue, we first use a convolution operation to align the channel dimensions and employ two parallel branches for average pooling and max pooling before fusing these branches. The TFF strategy is shown in Figure~\ref{fig:9}.

For combining features $N_i$ (where $i=3,4$), the feature $P_{i-1}$ of the layer is processed through a parallel structure of maximum pooling and average pooling, resulting in $P_{i-1}'$. The feature $PIG_i$ from the PIG module and $N_i$ are processed to obtain $PIG_i'$ and $N_i'$ via standard convolution. These features from the three branches are concatenated to obtain the final output $P_i$. For $i=2$ or $i=5$, the upper and lower branches are dropped,respectively. This approach fully integrates the advantages of both pooling methods, maximizing the extraction of background and texture information. Additionally, this strategy uses position information features from the PIG module as a priori knowledge to enhance the network's location awareness. This ensures the optimization of cross-scale interactions and location-aware features. Our model can detect dense and overlapping small targets within complex backgrounds with the help of the TFF module. The TFF module involves three inputs: the feature $N_i$ ($i=2,3,4,5$) from the intermediate neck network, the feature $P_{i-1}$ ($i=3,4,5$) from the neck network, and the feature $PIG_i$ ($i=2,3,4,5$) from the PIG module, as shown in Eq.~\ref{eq:15} below.

\begin{equation}
P_i=
\begin{cases}
Concat(N_i', PIG_i'), & i=2 \\
Concat(P_{i-1}', N_i', PIG_i'), & i=3,4 \\
Concat(P_{i-1}', N_i'), & i=5
\end{cases}
\label{eq:15}
\end{equation}

\begin{equation}
P_{i-1}'=Conv(Max(Conv(P_{i-1})),Avg(Conv(P_{i-1})))
\label{eq:16}
\end{equation}

\begin{equation}
N_i'=Conv(N_i)
\label{eq:17}
\end{equation}

\begin{equation}
PIG_i'=Conv(PIG_i)
\label{eq:18}
\end{equation}
$Max$ refers to maxpooling operation, and $Avg$ refers to average pooling opreation.

Finally, we integrate these modules to get a new model framework called BPIM. Specifically, the PIG module is placed after the final layer of the backbone network. Each output feature layer from the backbone network and each layer from the neck network serve as inputs to the BIG module. This arrangement captures dependencies of in-scale positional information and contextual boundary information in the feature maps.Furthermore, the CSF module employs 3D convolution to extract cross-scale interaction information from features at different layers. We combine this information with the small target detection layer $P_2$.Additionally, the TFF approach integrates features from $PIG_i$, $N_i$, and $P_i$ for efficient feature fusion. In the end, the AWF module adaptively fuses information across layers  at the neck network's output to address inconsistencies in cross-scale information. We have validated the effectiveness of our approach through experiments on three different datasets.

\section{Experiments and Analysis}

\subsection{Datasets}
To evaluate the effectiveness of our framework, we conducted experiments using three public datasets commonly used for small object detection, which are VisDrone2021\cite{visdrone}, DOTA1.0\cite{dota}, and WiderPerson\cite{widerperson}.

\textbf{VisDrone2021 Dataset} This dataset is designed for UAV vision tasks. It includes 288 video clips, 261,908 video frames, and 10,209 keyframe images. The dataset supports different scenarios, such as image and video object detection, single target and multi-target tracking. For image object detection, the dataset includes a total of 10 categories, the dataset is divided into 6471 images for training, 548 images for validation, 3190 images for testing.

\textbf{DOTA1.0 Dataset} The dataset includes 2,806 aerial images annotated by experts in aerial image interpretation, with a total of 188,282 instances classified into 15 categories. Due to the high resolution of the images, they are often cropped to increase the dataset size. For our experiments, we cropped the images to 1024 pixels using a step size of 200 pixels. The dataset is divided into 1411 images for the training set , 458 images for the validation set and 937 images for the test set.

\textbf{WiderPerson Dataset}This dataset is commonly used for outdoor dense pedestrian detection and contains 13,382 images with over 400,000 annotated instances, including five annotation types: persons, bicyclists, partially visible persons, crowds, and ignored regions. The dataset is divided into 8,000 images for training, 1,000 images for validation, and an unlabeled test set with 4382 images. As is common practice, we discarded the "ignored regions" category and unified the remaining four categories as "pedestrians" for training and validation.

\begin{table*}[ht]
\centering
\caption{ Dataset statistics}
\label{tab0}
\begin{tabular}{|l|c|c|c|c|c|c|}
\hline
\textbf{Dataset}& \textbf{Categories} & \textbf{Number of Images} & \textbf{Annotations ($\times10^3$)} & \textbf{Training Set} & \textbf{Validation Set} & \textbf{Test Set} \\ \hline
VisDrone2021\cite{visdrone}& 10& 10209& 343& 6471& 548& 3190\\ \hline
DOTA1.0\cite{dota}&15& 2806& 188& 1411& 458& 937\\ \hline
WiderPerson\cite{widerperson}& 5& 13382 & 400 & 8000& 1000 & 4382\\ \hline
\end{tabular}
\end{table*}

\subsection{Evaluation Metrics}
We aim to prove the robustness and effectiveness of our proposed method across different small object detection scenarios by conducting experiments on these datasets. 
We use $CIoU$ \cite{ciou}to capture the spatial relationship between the frames by considering the aspect ratio between the predicted and true frames. The $CIoU$ between the predicted bounding boxes $B$ and the ground truth bounding boxes $B_{gt}$ shows in Eq.~\ref{eq:19}:

\begin{equation}
CIoU(B, B_{gt}) = \text{IoU}(B, B_{gt}) - \frac{\rho^2}{c^2} - \alpha v
\label{eq:19}
\end{equation}

\begin{equation}
\alpha = \frac{v}{1 - \text{IoU}(B, B_{gt}) + v}
\label{eq:21}
\end{equation}

\begin{equation}
v = \frac{4}{\pi^2} \left( \arctan \left( \frac{w_{gt}}{h_{gt}} \right) - \arctan \left( \frac{w}{h} \right) \right)^2
\label{eq:20}
\end{equation}

The parameter $v$ in Eq.~\ref{eq:20} represents the normalization of the aspect ratio difference between the predicted and ground truth frames,  ranging from 0 to 1. In Eq.~\ref{eq:20}, $w$ and $h$ denote the width and height of the predicted frame, and $w_{gt}$ and $h_{gt}$ are the width and height of the ground truth frame. The offset factor $\alpha$ determines the relative importance of the aspect ratio term compared to the intersection over union (IoU) component in the loss calculation. $\rho$ is the distance between the centroids of the predicted and ground truth frames, while $c$ represents the length of the smallest enclosing rectangle that covers both frames to determine the diagonal length.
$CIoU$ evaluates the difference between the predicted and actual bounding boxes across multiple dimensions, demonstrating superior performance over traditional methods. Therefore, we calculate the confidence loss of the predicted box using 1-$CIoU$ between the predicted and labeled boxes,  assigning the confidence score for the predicted bounding box based on the $CIoU$ with the ground truth box. A higher $CIoU$ indicates a greater overlap with the ground truth, thus increasing the possibility that an object exists in this bounding box.

In summary, the evaluation metrics used in our experiments include mAP@.5, mAP@.5:.95, number of parameters, and computational load. mAP@.5 represents the mean Average Precision (mAP) at an IoU threshold of 0.5, while mAP@.5:.95 is the average of ten mAP values calculated at IoU thresholds ranging from 0.5 to 0.95, in increments of 0.05.

\subsection{Experimental Setup}
We use an NVIDIA RTX 3090 GPU, CUDA v12.0, Ubuntu 18.04 operating system, Python 3.8, and the PyTorch deep learning framework. Training was conducted using stochastic gradient descent (SGD) with an initial learning rate of 0.01, weight decay of 0.0005, and momentum of 0.9. Warmup training was performed during the first three epochs, with a warmup bias of 0.1 and momentum of 0.8. The batch size for all experiments was set to 8.

\subsection{Comparison results with baseline methods}
We compare the detection performance of BPIM on three datasets: VisDrone2021, DOTA1.0, and WiderPerson. We use YOLOv5n and YOLOv5l with a small object detection layer output $P_2$ as baseline models. 

In Table ~\ref{tab1}, BPIM obviously outperforms YOLOv5n-P2 in detection performance on VisDrone2021. BPIM can improve mAP@.5:.95 by 2.25\% and mAP@.5 by 2.72\% compared to YOLOv5n-P2. 
Furthermore, BPIM can enhance mAP@.5:.95 by 0.84\% and mAP@.5 by 2.45\% compared to YOLOv5l-P2.

\begin{table*}[hb]
\centering
\caption{Results on The VisDrone2021 Dataset. BPIM(YOLOv5n) indicates the proposed BPIM based on YOLOv5n framework, while BPIM(YOLOv5l) stands for the proposed BPIM based on YOLOv5l framework.}
\label{tab1}
\begin{tabular}{| l | c | c | c | c | c |}
\hline
Model & Resolution & mAP@.5:.95 & mAP@.5 & Parameters (M) & GFLOPs \\
\hline
YOLOv5n-P2 & 640$\times$640 & 16.29 & 30.38 & 1.781008 & 5.0 \\
\hline
\textbf{BPIM}(YOLOv5n) & 640$\times$640 & \underline{18.54} & \underline{33.10} & 2.819422 & 7.1 \\
\hline
YOLOv10n\cite{ref22} & 640$\times$640 & \textbf{19.80} & \textbf{33.60} & 2.698706 & 8.2 \\
\Xhline{0.5mm}
YOLOv5l-P2 & 640$\times$640 & 28.87 & 47.18 & 47.139572 & 127.2 \\
\hline
\textbf{BPIM}(YOLOv5l) & 640$\times$640 & \underline{29.71} & \textbf{49.63} & 56.802729 & 155.4 \\
\hline
YOLOv7\cite{ref20} & 640$\times$640 & \textbf{29.75} & \underline{49.58} & 37.250496 & 105.3 \\
\hline
YOLOv10l\cite{ref22} & 640$\times$640 & 28.10 & 45.40 & 25.733330 & 126.4 \\
\Xhline{0.5mm}
\textbf{BPIM}(YOLOv5l)   & 960$\times$960 & \underline{34.93} & 55.64 & 51.72  &  137.4 \\
\hline
\textbf{BPIM}(YOLOv5l)  & 1024$\times$1024 & \textbf{35.81} & \underline{57.35} & 51.72  &  145.1 \\
\hline
CEASC\cite{ref31} & 1333$\times$800 & 28.74 & 50.73 & - & 150.2 \\
\hline
CZ Det\cite{ref32} & 1500$\times$1500 & 33.21 & \textbf{58.34} & - & - \\
\hline
\end{tabular}
\end{table*}

Table ~\ref{tab2} compares the experimental results of our new framework with baseline methods on the DOTA1.0 dataset. When using YOLOv5n-P2 as baseline, BPIM can increase by 2.41\% in mAP@.5:.95 and 2.27\% in mAP@.5.
BPIM can heighten by 1.35\% in mAP@.5:.95 and 1.85\% in mAP@.5 compared to the baseline YOLOv5l-P2.

\begin{table*}[ht]
\centering
\caption{Results on The DOTA1.0 Dataset.BPIM(YOLOv5n) indicates the proposed BPIM based on YOLOv5n framework, while BPIM(YOLOv5l) stands for the proposed BPIM based on YOLOv5l framework.}
\label{tab2}
\begin{tabular}{| l | c | c | c | c | c |}
\hline
Model & Resolution & mAP@.5:.95 & mAP@.5 & Parameters (M) & GFLOPs \\
\hline
YOLOv5n-P2 & 640$\times$640 & 40.42 & 66.70 & 1.78 & 4.8 \\
\hline
\textbf{BPIM}(YOLOv5n) & 640$\times$640 & \underline{42.83} & \textbf{68.97} & 2.83 & 7.1 \\
\hline
YOLOv10n\cite{ref22} & 640$\times$640 & \textbf{45.00} & \underline{67.70} & 2.70 & 8.3 \\
\Xhline{0.5mm}
YOLOv5l-P2 & 640$\times$640 & 52.16 & 76.82 & 47.17 & 127.4 \\
\hline
\textbf{BPIM}(YOLOv5l) & 640$\times$640 & \textbf{53.51} & \textbf{78.67} & 56.83 & 155.7 \\
\hline
YOLOv7\cite{ref20} & 640$\times$640 & \underline{53.34} & \underline{78.53} & 37.15 & 105.1 \\
\hline
YOLOv10l\cite{ref22} & 640$\times$640 & 49.70 & 72.40 & 25.74 & 126.4 \\
\hline
FCOSR-L\cite{ref33} & 1024$\times$1024 & - & 77.4 & 89.64 & 445.7 \\
\hline
PP-YOLOE-R-l\cite{ref34} & 1024$\times$1024 & - & 78.1 & 53.29 & 281.6 \\
\hline
\end{tabular}
\end{table*}

Table ~\ref{tab3} shows the result on the WiderPerson dataset. Wether the baseline methods are YOLOv5n-P2 or YOLOv5l-P2, the proposed BPIM both have a great improvement comparing with them. That indicates BPIM can sufficiently use the position and boundary information, which boost by the PIG moduel, the BIG module, the TFF module, the AWF module and the CSF module.
These results show that BPIM significantly reinforces the detection ability of the position and edge information, and the proposed BPIM can provide a clear performance boost for both YOLOv5n and YOLOv5l.

\begin{table*}[ht]
\centering
\caption{Results on The WiderPerson Dataset.BPIM(YOLOv5n) indicates the proposed BPIM based on YOLOv5n framework, while BPIM(YOLOv5l) stands for the proposed BPIM based on YOLOv5l framework.}
\label{tab3}
\begin{tabular}{| l | c | c | c | c | c |}
\hline
Model & Resolution & mAP@.5:.95 & mAP@.5 & Parameters (M) & GFLOPs \\
\hline
YOLOv5n-P2 & 640$\times$640 & 57.46 & 87.60 & 1.760680 & 4.8 \\
\hline
\textbf{BPIM}(YOLOv5n) & 640$\times$640 & \underline{59.95} & \textbf{89.00} & 2.806354 & 7.0 \\
\hline
YOLOv10n\cite{ref22} & 640$\times$640 & \textbf{63.10} & \underline{88.60} & 2.694806 & 8.2 \\
\Xhline{0.5mm}
YOLOv5l-P2 & 640$\times$640 & 64.01 & 90.16 & 47.087624 & 126.8 \\
\hline
\textbf{BPIM}(YOLOv5l) & 640$\times$640 & \underline{64.81} & \textbf{92.12} & 56.802631 & 155.3 \\
\hline
YOLOv7\cite{ref20} & 640$\times$640 & 64.79 & \underline{92.10} & 37.112573 & 105.0 \\
\hline
YOLOv10l\cite{ref22} & 640$\times$640 & \textbf{66.60} & 91.50 & 25.717910 & 126.3 \\
\hline
IterDet\cite{ref35} & 1024$\times$1024 & - & 91.94 & - & - \\
\hline
MSAGNet\cite{ref36} & 1024$\times$1024 & - & 92.11 & - & - \\
\hline
\end{tabular}
\end{table*}

\subsection{Comparison results with the state-of-the-art methods}

To evaluate the effectiveness of the proposed BPIM, we compared BPIM with the state-of-the-art methods, including YOLOv7\cite{ref20}, YOLOv10\cite{ref22}, CEASC\cite{ref31}, CZ Det\cite{ref32}, FCOSR-L\cite{ref33}, PP-YOLOE-R-l\cite{ref34}, IterDet\cite{ref35} and MSAGNet\cite{ref36}.

In Table ~\ref{tab1}, the proposed BPIM not only approximates to the highest detection accuracy of YOLOv10n on the VisDrone2021 dataset, but also requires less computation than YOLOv10n. This indicates that the proposed BPIM has lower configuration requirements for devices. Besides that, the performance of BPIM surpasses that of YOLOv10l. When the resolution of images increasing, the proposed BPIM can obtain the promising results compared with CEASC\cite{ref31} and CZ Det\cite{ref32}.

In Table ~\ref{tab2}, BPIM achieves the best performance comparing to YOLOv10l on the DOTA1.0 dataset, while BPIM performs better than YOLOv10n with a lower computational loading under the common metric index (for example mAP@.5). Although FCOSR-L\cite{ref33} and PP-YOLOE-R-l\cite{ref34} have the better performance in the high resolution of image, the proposed BPIM outperforms them in the low-resolution images.

In Table ~\ref{tab3}, the similar situation in the DOTA1.0 dataset occurs in the WiderPerson datasets. The proposed BPIM achieves the best performance in the common metric index. Even if the resolution is vital for small object detection, the performance of the proposed BPIM in low-resolution images surpasses that of IterDet\cite{ref35} and MSAGNet\cite{ref36} in high-resolution images. It shows that the proposed BPIM can effectively balance model parameters and computational load, resulting in a favorable trade-off among detection performance, model size, and computational efficiency. Therefore, the BPIM framework is well-suited for UAV aerial target detection.


\subsection{Ablation Study}
In order to illustrate the effectiveness of the modules of our proposed model, we further conducted ablation experiments on the experimental dataset. 
We focused on evaluating the BIG module and the AWF module in adaptive weight fusion with boundary strategy. Table \ref{tab6} presents the comparative performance results of the BPIM model components across three experimental datasets. Furthermore, the integration of PIG and CSF into the YOLOv5n model can strengthen the ability to capture position and cross-scale information. The proposed BPIM can fully utilize edge position after Introducing these modules, resulting in 1\% to 3\% in mAP@.5:.95 and mAP@.5. The final results demonstrate that BPIM can enhance boundary perception, capture contextual information features and positional distance dependencies of feature maps. The BPIM model pays attention to boundary ,position ,and cross-scale information, achieving the best performance after combining all the modules.

\begin{table*}[htbp]
\centering
\caption{Ablation Study Results of AWFB Module on Different Datasets}
\label{tab6}
\begin{tabular}{| l | c | c | c | c | c | c | c | c | c | c |}
\hline
Datasets & YOLOv5n-P2 & BIG & AWF & PIG & CSF & mAP@.5:.95 & mAP@.5 & Parameters (M) & GFLOPs \\
\hline
\multirow{4}{*}{\raisebox{-5.5\height}{VisDrone2021}}
 & \checkmark &  &  &  &  &16.29 & 30.38 & 1.781008 & 5.0 \\
 & \checkmark & \checkmark &  &  &  & 17.41 & 31.70 & 1.863783 & 5.4 \\
 & \checkmark &  & \checkmark &  &  & 17.31 & 31.56 & 1.967460 & 5.3 \\
 & \checkmark & \checkmark & \checkmark &  &  &  17.72 & 32.02 & 2.060478 & 6.0 \\
 & \checkmark &  &  & \checkmark &  & 16.59 & 30.78 & 1.784324 & 5.1 \\
 & \checkmark &  &  &  & \checkmark & 16.63 & 30.61 & 2.450064 & 5.4 \\
 & \checkmark &  &  & \checkmark & \checkmark & 16.87 & 31.28 & 2.517476 & 5.8 \\
 & \checkmark & \checkmark & \checkmark & \checkmark &\checkmark  & \textbf{18.54} & \textbf{33.10} & 2.819422 & 7.1 \\
\hline
\multirow{4}{*}{\raisebox{-5.5\height}{DOTA1.0}}
 & \checkmark &  &  &  &  &40.42 & 66.70 & 1.776032 & 4.8 \\
 & \checkmark & \checkmark &  &  &  & 41.06 & 67.47 & 1.868726 & 5.4 \\
 & \checkmark &  & \checkmark &  &  & 42.57 & 67.95 & 2.009016 & 5.4 \\
& \checkmark & \checkmark & \checkmark &  &  &43.15 & 68.42 & 2.067738 & 6.1 \\
 & \checkmark &  &  & \checkmark &   & 41.01 & 67.35 & 1.791584 & 5.1 \\
 & \checkmark &  &  &  & \checkmark & 40.91 & 67.31 & 2.444480 & 5.3 \\
 & \checkmark &  &  & \checkmark & \checkmark & 41.44 & 68.21 & 2.524736 & 5.8 \\
 & \checkmark & \checkmark & \checkmark & \checkmark &\checkmark  & \textbf{42.83} & \textbf{68.97} & 2.826682 & 7.1 \\
\hline
\multirow{4}{*}{\raisebox{-5.5\height}{WiderPerson}}
& \checkmark &  &  &  &  & 57.46 & 87.60 & 1.760680 & 4.9 \\
& \checkmark & \checkmark &  &  &  & 58.94 & 88.40 & 1.841938 & 5.3 \\
& \checkmark &  & \checkmark &  &  & 58.73 & 88.43 & 1.954338 & 5.2 \\
& \checkmark & \checkmark & \checkmark &  &  & 59.67 & 88.84 & 2.047410 & 5.9 \\
 & \checkmark &  &  & \checkmark &  & 57.58 & 87.88 & 1.771256 & 5.0 \\
 & \checkmark &  &  &  & \checkmark & 57.74 & 87.94 & 2.429736 & 5.4 \\
 & \checkmark &  &  & \checkmark & \checkmark & 59.09 & 88.59 & 2.504408 & 5.7 \\
 & \checkmark & \checkmark & \checkmark & \checkmark &\checkmark  & \textbf{59.95} & \textbf{89.00} & 2.806354 & 7.0 \\
\hline
\end{tabular}
\end{table*}

As a result, BPIM effectively balances the accuracy with computational complexity and parameters, achieving the best performance without significantly increasing computational load compared to the state-of-the-art methods. When the experimental images have high resolution, BPIM outperforms existing detection methods. In the ablation experiments, the effectiveness of the model modules is clearly demonstrated. All modules significantly reinforce small target detection in the baseline model.

\subsection{Visualization Analysis}
\begin{figure*}[ht]
\centering
\includegraphics[width=0.8\linewidth]{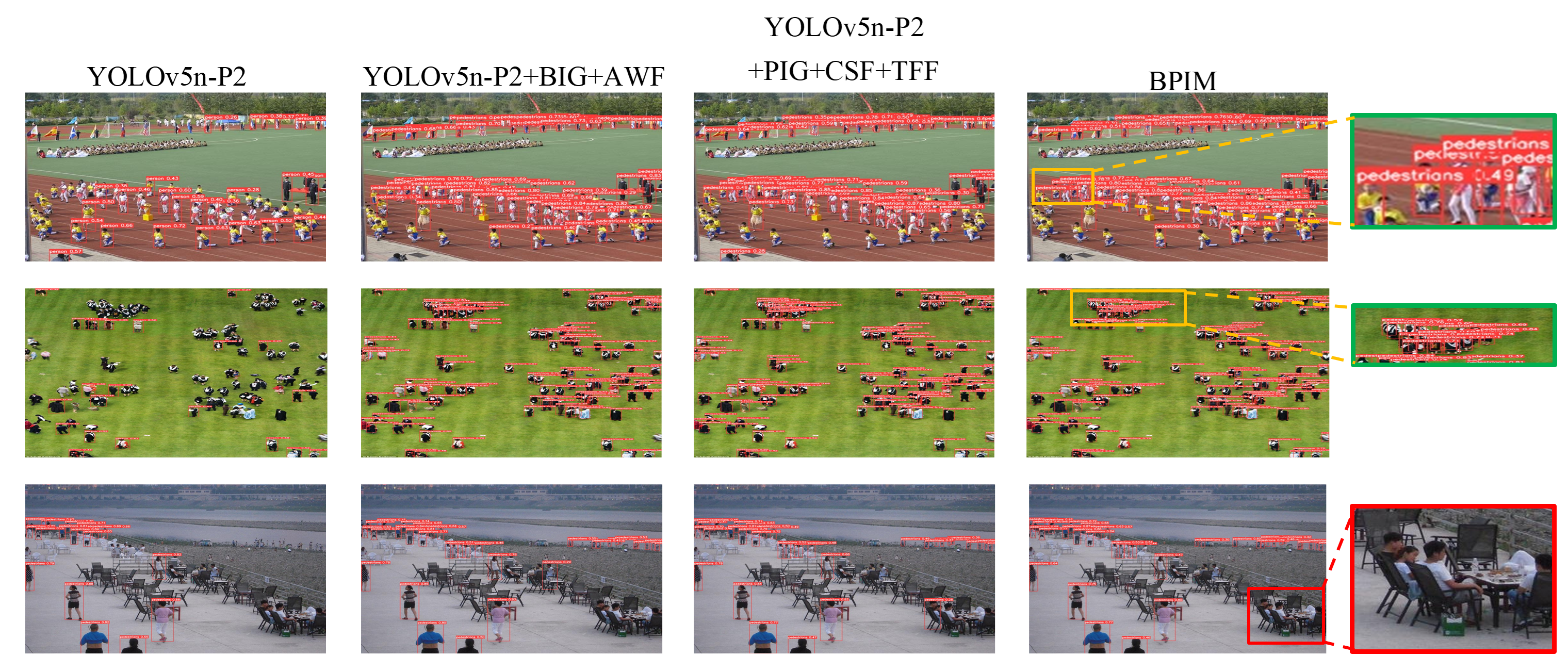}
\caption{\raggedright The detection visualization on the WiderPerson dataset.The first column of figures shows the detection effect of the YOLOv5n-P2, the second and third columns reveal the detection effect of adaptive weight fusion with boundary strategy based on YOLOv5n-P2(YOLOv5n-P2+BIG+AWF) and cross-scale position fusion strategy based on YOLOv5n-P2(YOLOv5n-P2+PIG+CSF+TFF),the forth column presents the detection effect of the proposed BPIM, the last column demonstrate the details of the detection effect.}
\label{fig:12}
\end{figure*}

To intuitively evaluate the small object detection capability of the different strategies, which are BIG and AWF modules for adaptive weight fusion with boundary and the PIG, CSF and TFF module for cross-scale position fusion, we show the detection visualization comparing the proposed BPIM and the combined modules with the YOLOv5n-P2. The selected detection images are sourced from the VisDrone2021, DOTA1.0, and WiderPerson datasets. By visually comparing the detection results on these datasets, we can clearly observe how BPIM performs in real-world scenarios, particularly in detecting small and dense target instances. This visualization illustrates the practical advantages and improvements of our proposed method compared to YOLOv5n-P2. However, the proposed BPIM demonstrates missing detection in some situations, particularly when objects are heavily occluded or in dense clutter. For example, in the third row of Figure ~\ref{fig:12}, a missing detection is shown in this case.

\begin{figure*}[ht]
\centering
\includegraphics[width=0.8\linewidth]{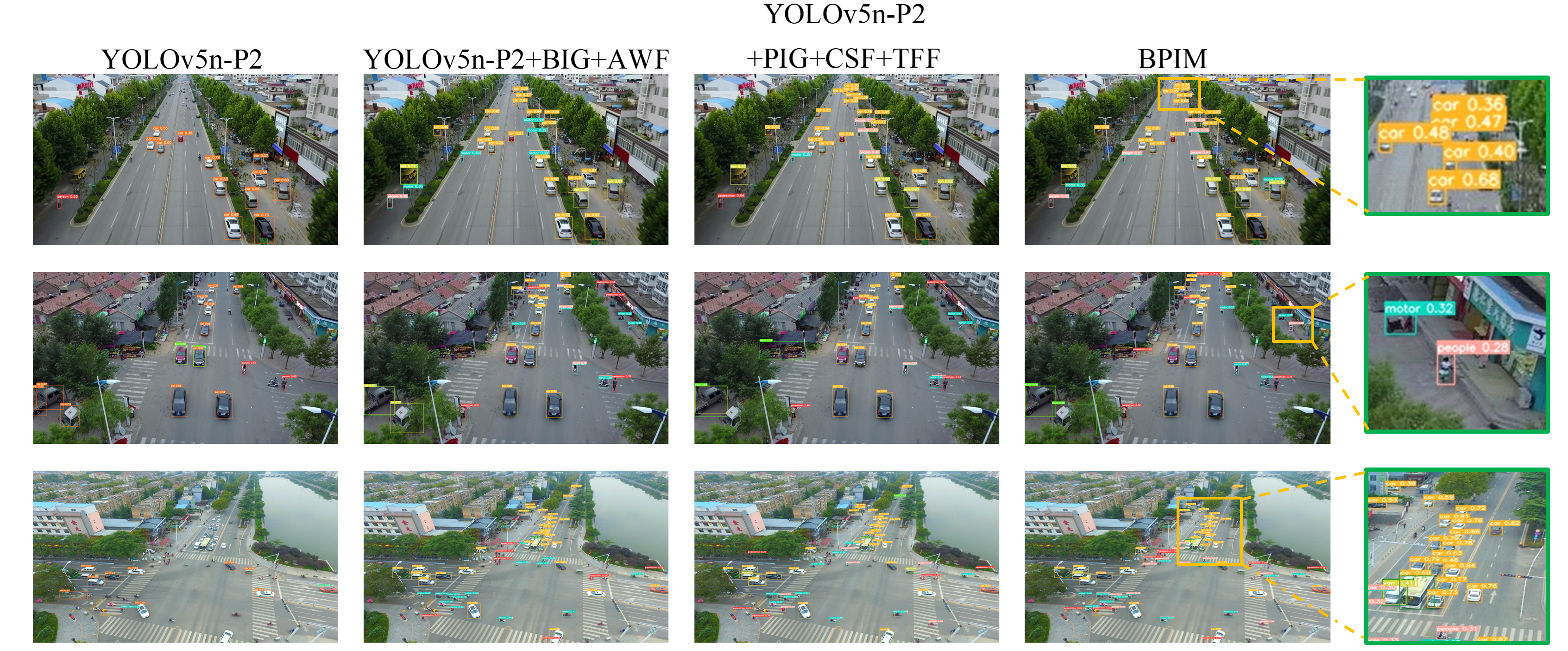}
\caption{\raggedright The detection visualization on the VisDrone2021 dataset.The first column of figures shows the detection effect of the YOLOv5n-P2, the second and third columns reveal the detection effect of adaptive weight fusion with boundary strategy based on YOLOv5n-P2(YOLOv5n-P2+BIG+AWF) and cross-scale position fusion strategy based on YOLOv5n-P2(YOLOv5n-P2+PIG+CSF+TFF),the forth column presents the detection effect of the proposed BPIM, the last column demonstrate the details of the detection effect.}
\label{fig:10}
\end{figure*}

\begin{figure*}[ht]
\centering
\includegraphics[width=0.8\linewidth]{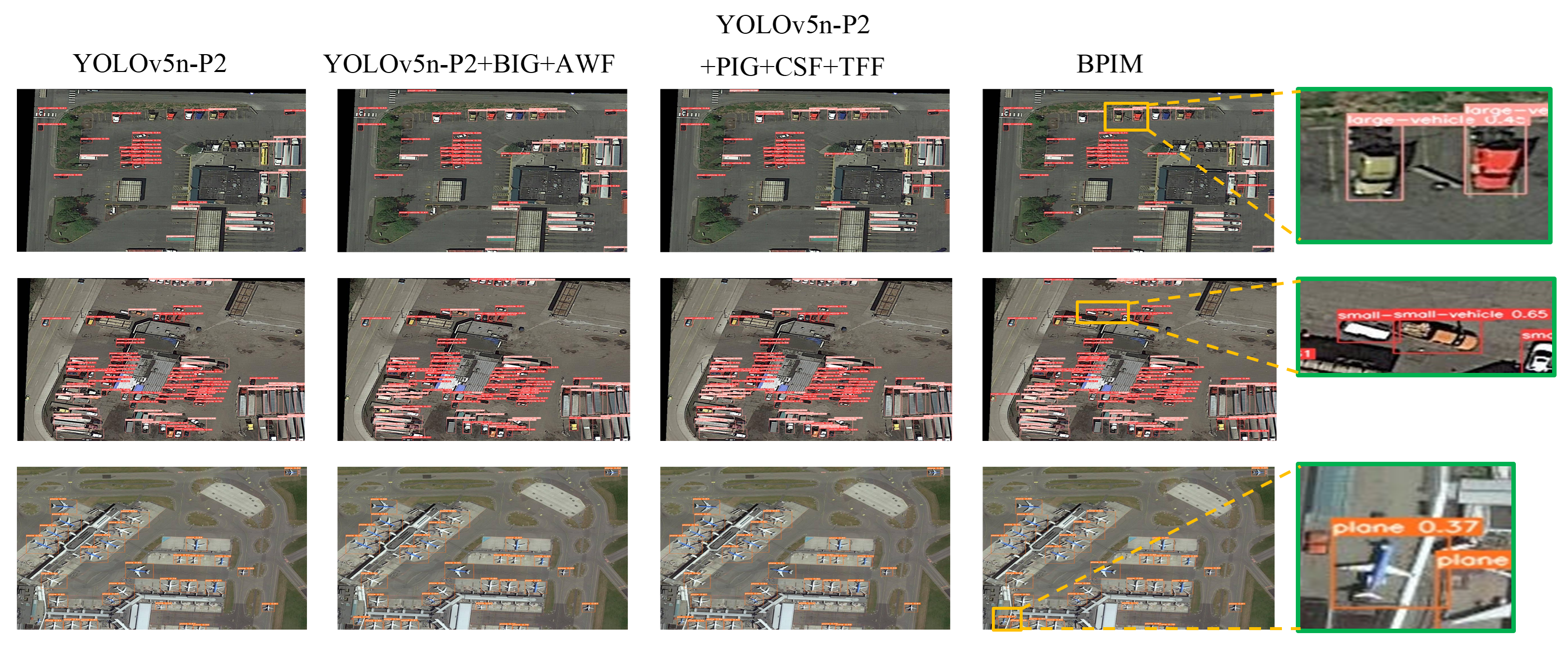}
\caption{\raggedright The detection visualization on the DOTA1.0 dataset.The first column of figures shows the detection effect of the YOLOv5n-P2, the second and third columns reveal the detection effect of adaptive weight fusion with boundary strategy based on YOLOv5n-P2(YOLOv5n-P2+BIG+AWF) and cross-scale position fusion strategy based on YOLOv5n-P2(YOLOv5n-P2+PIG+CSF+TFF),the forth column presents the detection effect of the proposed BPIM, the last column demonstrate the details of the detection effect.}
\label{fig:11}
\end{figure*}

\section{Conclusion}

To address the imbalance scale and the blurred edges in aerial small object detection, we present the BPIM framework, which improves small object detection performance through two strategies: adaptive weight fusion with boundary and cross-scale position fusion. Firstly,  adaptive weight fusion with boundary can be implemented by the BIG and AWF modules. The BIG module can extract the detailed boundary information from the output feature maps of each layer in the backbone network, while the AWF modules adaptively learns fusion weights for different layer features, enhancing the detection of target instances. Secondly, cross-scale position fusion include the PIG, CSF and TFF modules. The PIG module captures interrelationships and long-range dependencies between pixel positions at the end of the backbone network, the CSF module utilizes 3D convolution along the scale transformation direction for cross-scale feature extraction, and the TFF module integrates the PIG module with middle-layer and output-layer features of the backbone network to grasp interaction information between features at various scales, thereby improving feature representation. The experimental results show that BPIM outperforms the state-of-the-art methods in both detection accuracy and computational efficiency. However, the proposed BPIM does have certain limitations, namely the challenges of complex occlusion and computational load for embedded systems. Future work will focus on developing lightweight models to reduce parameters and computational demands without sacrificing performance in video tracking.

\section{Acknowledgments}
The authors would like to thank the anonymous reviewers for their insightful comments that can improve the quality of this paper. This work was supported by NSFC (Program No. 61771386), Key Research and Development Program of Shaanxi (Program No. 2020SF-359) and Natural Science Basic Research Plan in Shaanxi Province of China (Program No. 2021JM-340).

\bibliographystyle{IEEEtran}  
\bibliography{BPIM_refs}      
\end{document}